\documentclass{article}

    \PassOptionsToPackage{numbers, compress}{natbib}
    \usepackage[preprint]{neurips_2024}

\usepackage[utf8]{inputenc} % allow utf-8 input
\usepackage[T1]{fontenc}    % use 8-bit T1 fonts
\usepackage{hyperref}       % hyperlinks
\usepackage{url}            % simple URL typesetting
\usepackage{booktabs}       % professional-quality tables
\usepackage{amsfonts}       % blackboard math symbols
\usepackage{nicefrac}       % compact symbols for 1/2, etc.
\usepackage{microtype}      % microtypography
\usepackage{xcolor}         % colors
\usepackage{multirow}
\usepackage{colortbl}
\usepackage{algorithm}
\usepackage{algpseudocode}
\usepackage{graphicx}
\usepackage{booktabs}
\usepackage{amsmath}
\usepackage{cleveref}
\crefname{figure}{Fig.}{Figs.}
\crefname{appendix}{Appendix}{Appendices}
\crefname{table}{Table}{Tables}
\crefformat{equation}{#2Eq.~#1#3}

\title{LoRA-Composer: Leveraging Low-Rank Adaptation for Multi-Concept Customization in Training-Free Diffusion Models}

\author{%
Yang Yang$^{1}$, 
Wen Wang$^{1}$, 
Liang Peng$^{2}$, 
Chaotian Song$^{3}$, 
Yao Chen$^{3}$, 
Hengjia Li$^{1}$, \\
\textbf{Xiaolong Yang$^{4}$, 
Qinglin Lu$^{4}$, 
Deng Cai$^{1}$, 
Boxi Wu$^{3}$, 
Wei Liu$^{4}$}
\\\\
 $^1$State Key Lab of CAD\&CG, Zhejiang University, \quad$^2$Fabu Inc. \\
 $^3$The School of Software Technology, Zhejiang University, \quad$^4$Tencent Inc.
}

\begin{document}

\maketitle

\begin{abstract}
Customization generation techniques have significantly advanced the synthesis of specific concepts across varied contexts. Multi-concept customization emerges as the challenging task within this domain. Existing approaches often rely on training a fusion matrix of multiple Low-Rank Adaptations (LoRAs) to merge various concepts into a single image. However, we identify this straightforward method faces two major challenges: 1) concept confusion, where the model struggles to preserve distinct individual characteristics, and 2) concept vanishing, where the model fails to generate the intended subjects. To address these issues, we introduce LoRA-Composer, a training-free framework designed for seamlessly integrating multiple LoRAs, thereby enhancing the harmony among different concepts within generated images.
LoRA-Composer addresses concept vanishing through concept injection constraints, enhancing concept visibility via an expanded cross-attention mechanism. To combat concept confusion, concept isolation constraints are introduced, refining the self-attention computation. Furthermore, latent re-initialization is proposed to effectively stimulate concept-specific latent within designated regions. Our extensive testing showcases a notable enhancement in LoRA-Composer's performance compared to standard baselines, especially when eliminating the image-based conditions like canny edge or pose estimations. Code is released at \url{https://github.com/Young98CN/LoRA_Composer}
% Codes will be made available.
\end{abstract}

\section{Introduction}
\label{sec:intro}

% The advancement of d
Diffusion models~\cite{rombach2021highresolution} have significantly advanced the field of image generation, particularly in creating images that adhere to user-specific concepts.
% , thereby enriching the landscape of image synthesis. 
% This progress plays an important role in 
The progress made in customization models~\cite{gu2023mixofshow, kumari2022multiconcept, shah2023ziplora, ruiz2022dreambooth, yang2023paint, chen2023anydoor} play an important role in enriching the landscape of image synthesis. 
As technologies for single concept customization evolve, users are presented with various methods to personalize content, ranging from fine-tuning U-Net~\cite{ruiz2022dreambooth,kumari2022multiconcept}, modifying text embeddings~\cite{gal2022image,cones2}, to leveraging Low-Rank Adaptations (LoRA)~\cite{hu2021lora}. LoRA is a versatile, plug-and-play module that enables users to customize their models to generate diverse and lifelike personal images. Its adaptability and accuracy in image generation have established LoRA as a preferred method for customization tasks, significantly influencing how the community approaches the creation of tailored visual content.

\begin{figure}[!tb]
  \centering
  \includegraphics[width=0.98\textwidth]{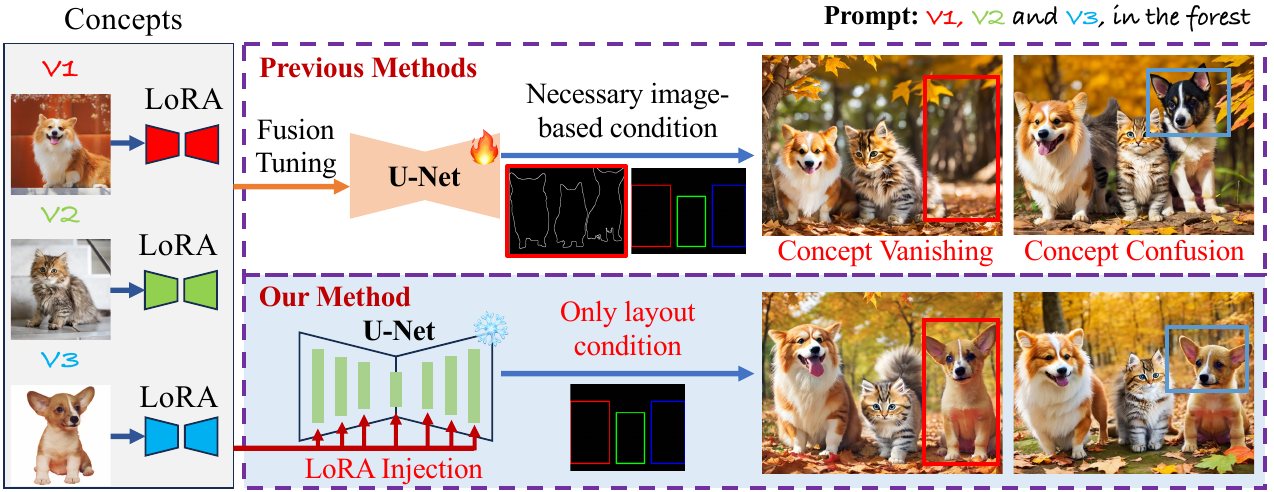}
  \vspace{-3mm}
  \caption{Our method distinguishes itself from Mix-of-Show~\cite{gu2023mixofshow} by eliminating the image-based conditions and the requirement to train a LoRA fusion matrix. Furthermore, we highlight the limitations of Mix-of-Show through the demonstration of failure cases. In the top row, we illustrate two key issues: concept vanishing, marked by the absence of intended concepts in the image, and concept confusion, where the model mistakenly merges and confuses distinct concepts.
  }
  \label{fig:intro}
  \vspace{-6mm}
\end{figure}

While LoRA excels in single-concept customization, its application to emerging multi-concept customization tasks presents challenges. Recent developments have explored the integration of multiple-concept LoRAs to infuse images with diverse concepts via fusion tuning~\cite{gu2023mixofshow, wang2023autostory}. However, as illustrated in \cref{fig:intro}, these integration strategies often necessitate a variety of conditions, including textual and image-based inputs~\cite{zhang2023adding} (such as human pose estimations, and canny edge detection). This approach introduces constraints on variation and flexibility. 
Furthermore, in the process of combining multiple LoRAs, prior research~\cite{gu2023mixofshow, wang2023autostory,smith2023c_lora} focused on training a fusion ratio matrix, which aims to optimally weigh individual LoRAs. However, adjusting LoRA weights in this manner can exacerbate two problems: 1) concept vanishing, where the concept fails to be injected into the figure; and 2) concept confusion, where the model struggles to associate attributes with subjects or fails to capture distinct concept characteristics. Examples illustrating the issues mentioned are displayed in the top row of \cref{fig:intro}, showcasing outputs from the representative method Mix-of-Show~\cite{gu2023mixofshow}. The left column shows a clear case of concept vanishing, where the model fails to generate one of the dogs (highlighted within the red box). The right column highlights issues with incorrect attribute binding, such as the dog's color being mistaken (indicated within the blue box).
    
To overcome existing challenges, we introduce LoRA-Composer, a training-free framework that enables the synthesis of images with multiple concept LoRAs, utilizing textual and layout cues. LoRA-Composer encompasses three principal components: concept injection constraints, concept isolation constraints, and latent re-initialization. 
The concept injection constraints introduce a novel cross-attention mechanism, consisting of 1) region-aware LoRA injection, which injects concept-specific LoRA features into designated regions through cross-attention, facilitating the seamless integration of multiple LoRAs within the diffusion model framework without the need for fusion fine-tune. 2) concept enhancement constraints, which guide the refinement of latents to accentuate concepts in user-specified regions. These strategies enable the model to concentrate on areas designated for concept insertion, effectively mitigating the issue of concept vanishing.
The concept isolation constraints specifically address the issue of concept fusion by concentrating on self-attention. They implement restrictions to guarantee that each concept maintains its unique characteristics. 
% This plays a crucial role in guiding the denoising process within the latent space, maintaining alignment with each concept's unique features and corresponding textual descriptions.
Due to traditional single-concept LoRA training typically without layout conditions, which can be restrictive for localized area generation. We propose re-initializing the latent vector to establish a more accurate prior, directing the model's focus on specific areas of the image.

We rigorously test our LoRA-Composer across a broad spectrum of multi-concept customization scenarios, including categories such as animals, characters, and scenic backgrounds. Our approach displayed a strong performance compared to existing benchmarks through comprehensive qualitative and quantitative assessments. 
% Notably, our approach facilitates the customization of numerous subjects and all image elements (both background and foreground) in a variety of styles, such as anime and realistic, as demonstrated in \cref{fig: main_res}(a). Moreover, our method can manipulate interactions and attributes by textural prompt, demonstrated in \cref{fig: main_res}(b). Furthermore, it flexibly generates images under diverse conditions, incorporating constraints like edge detection or pose estimation, detailed in \cref{fig: main_res}(c).
In summary, our contributions are as follows:
\begin{itemize}
    % \item We explore the challenges in multi-concept customization, specifically concept vanishing and confusion, and the limitations of LoRA-based models, which include the need for numerous conditions.
    \item We propose a training-free model for integrating multiple LoRAs called LoRA-Composer. It requires only easily accessible conditions: layout and textual prompts. This approach simplifies the process of multi-concept customization, enhancing convenience.
    % combining various concepts into a cohesive image, enhancing the convenience for users.
    \item To tackle concept vanishing and confusion, we implement concept injection constraints and concept isolation constraints. These strategies enhance the attention mechanism in U-Net, enabling the model to concentrate on the characteristics of individual concepts and prevent interference from the background or other concepts.
    % These approaches are designed to guide the latent update process using attention maps.
    \item We propose latent re-initialization to obtain a better prior enhancing the model's capability to focus on specific image sections.
    \item Our extensive evaluations reveal that our method exceeds baseline performance, particularly in scenarios that eliminate image-based conditions.
\end{itemize}

\vspace{-2mm}
\section{Related Work}
\vspace{-1mm}

% \subsection{Diffusion Model}

% \subsection{Customized Concept Generation}

\subsection{Controllable Image Generation}
% - ControlNet, T2I-Adaptor, ...

Diffusion models~\cite{ho2020denoising,sohl2015deep} trained on large-scale text-to-image datasets, like DALLE-2~\cite{ramesh2022hierarchical}, Imagen~\cite{saharia2022photorealistic}, and Stable Diffusion~\cite{rombach2021highresolution}, SDXL~\cite{podell2023sdxl} can produce text-aligned and diverse images in unprecedented high quality.
To further support image generation from fine-grained spatial conditions, like sketches, human keypoints, semantic maps, \textit{etc.}, ControlNet~\cite{zhang2023adding} finetunes a trainable copy of the pre-trained U-Net and connects the new layers and original U-Net weights with zero convolutions. 
A similar work, T2I-Adaptor~\cite{mou2023t2i}, finetunes lightweight adaptors for conditional generation from spatial conditions. 
Differently, GLIGEN~\cite{li2023gligen} considers controllable generation with sparse box layout conditions and injects a gated self-attention for fine-tuning.
% 
% While these works rely on self-collected datasets of condition-image pairs for fine-tuning, 
Recent works~\cite{xie2023boxdiff,phung2023grounded} seek to explore test-time optimization for zero-shot controllable generation. 
For example, both BoxDiff~\cite{xie2023boxdiff} and Attention Refocusing~\cite{phung2023grounded} achieve zero-shot layout conditioned generation, by maximizing the attention weights between the features inside the box and its corresponding text description, while discouraging the latent features outside the box from attending to the text.

\vspace{-3mm}
\subsection{Multi-Concept Customization}
\vspace{-2mm}

Concept customization aims at generating concepts specified by a few input images.
% 
% While significant progress has been made in generating a single custom concept~\cite{gal2022image,ruiz2022dreambooth,li2023blipdiffusion,voynov2023p+,alaluf2023neural}, the customized generation of multiple concepts remains challenging.
While significant progress has been made in generating a single custom concept~\cite{ruiz2022dreambooth,gal2022image,tewel2023key,shi2023instantbooth,wei2023elite,han2023svdiff,nichol2021improved, li2023blipdiffusion, ruiz2023hyperdreambooth}, the customized generation of multiple concepts remains challenging.
% 
% - Inpaiting: PBE, Anydoor
% 
% Different from the above work that focuses on single-concept customization, several recent works~\cite{kumari2022multiconcept,liu2023cones,liu2023cones2,avrahami2023break,xiao2023fastcomposer} tackle the more challenging generic setting of multiple concepts. 
% 
A pioneer work, Custom Diffusion~\cite{kumari2022multiconcept}, jointly finetunes multiple concept images for customization.
Cones series~\cite{liu2023cones,liu2023cones2} finds concept-related neurons in pre-trained diffusion models for multi-concept customization. 
% 
% Cone 2~\cite{liu2023cones2} further incorporates residual token embedding and layout conditioning for better generation quality.
% 
To accelerate the customized generation, FastComposer~\cite{xiao2023fastcomposer} finetunes diffusion model on massive data to take subject embedding as input and generate the composed image of multiple concepts.
Similarly, Paint-by-Example~\cite{yang2023paint} and AnyDoor~\cite{chen2023anydoor} are trained on a significant amount of images and can achieve multi-concept generation through image inpainting.
Considering the widespread utilization of LoRA for customization
% in the community
, several recent works~\cite{gu2023mixofshow,wang2023autostory,shah2023ziplora,zhong2024multi} seek to achieve multi-concept customization by combining multiple LoRA weights of individual concepts.
For example, Mix-of-show~\cite{gu2023mixofshow} proposes gradient fusion to train a composed LoRA weight that mimics the prediction of individual LoRAs. It further leverages T2I-Adaptor~\cite{mou2023t2i} and sketches or key points for final generation.
A concurrent work~\cite{zhong2024multi} proposes two variants, namely LoRA Swich and LoRA Composite, to realize LoRA merge during decoding. The former uses multiple LoRA weights sequentially, while the latter averages the latent obtained from different LoRA weights. 
% 
% By contrast, we merge the latent of different LoRAs based on simple layouts, which effectively alleviates the spatial conflicts when multiple LoRA weights are applied to the same locations.
% concept confusion or vanishing problem.
However, they focus on the combination of a single foreground and a single background LoRA weights. 
By contrast, we tackle the more challenging task of customizing multiple foreground characters, facing issues of concept confusion and vanishing.

Probably the most similar work to ours is Mix-of-Show, but we emphasize the following differences. 
Firstly, Mix-of-Show requires repeated gradient fusion training for each combination of multiple concepts, while we achieve this on the fly without retraining the LoRA weight. 
Secondly, Mix-of-Show requires additional image-based conditions, like sketches and keypoints, as input for high-quality image generation, which could be difficult to obtain.

\vspace{-2mm}
\section{Method}
\begin{figure}[t]
  \centering
  \includegraphics[width=\textwidth]{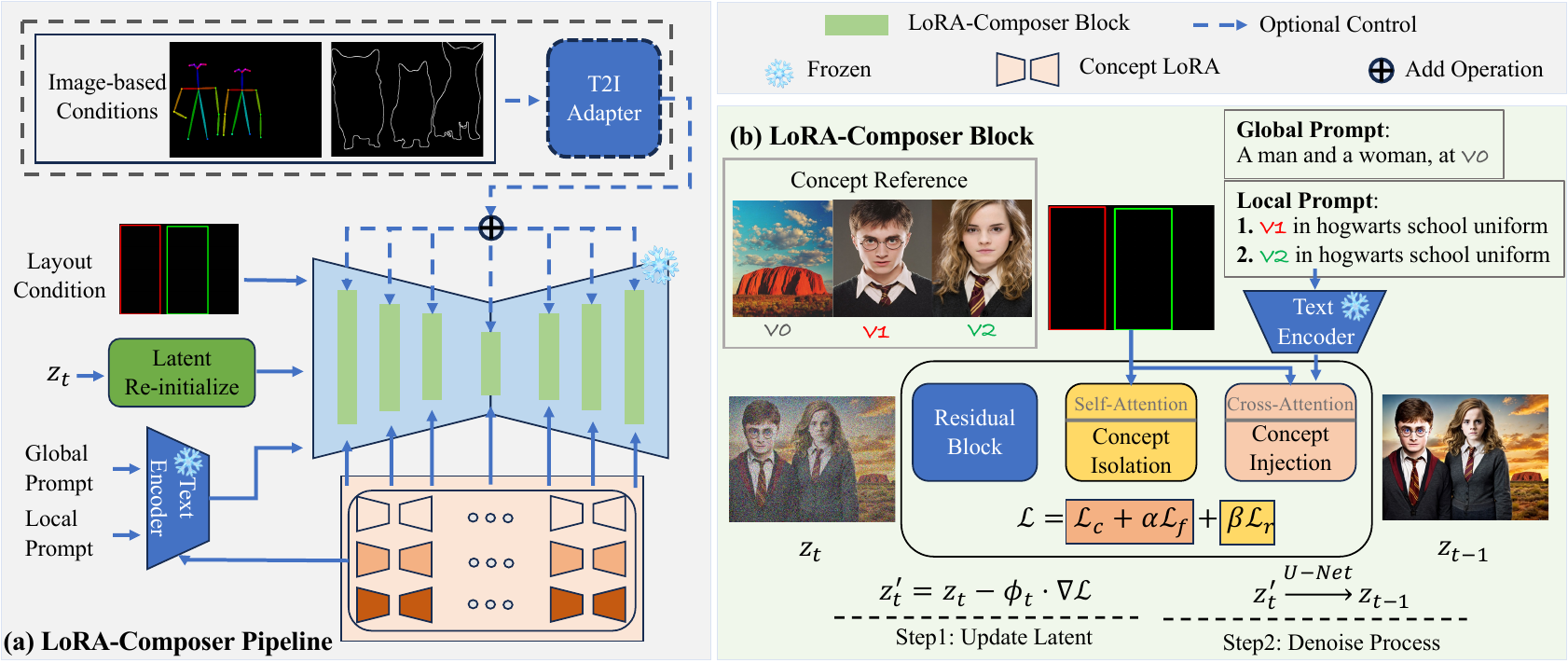}
  \vspace{-5mm}
  \caption{(a) LoRA-Composer utilizes textual, layout, and image-based conditions (optional) to integrate multiple LoRAs. (b) Modifications to the U-Net in LoRA-Composer Block include concept isolation in self-attention and concept injection in cross-attention. At timestep $t$, $z_t$ is first refined via $\mathcal{L}$ to ensure appearance consistency and prevent feature leakage, followed by the denoising process.
  }
  \label{fig:method}
  \vspace{-5mm}
\end{figure}

In this section, 
% we commence with a concise overview of diffusion models and ED-LoRA in \cref{sec:Preliminary}. Following this, 
we introduce our innovative LoRA-Composer approach in \cref{sec: pipeline overview}, with the pipeline depicted in \cref{fig:method}(a). The key point is augmenting the scalability of LoRA through the utilization of the LoRA-Composer Block, as illustrated in \cref{fig:method}(b). We then delve into the specifics of the two primary components of the LoRA-Composer Block, outlined in \cref{sec: concept injection} and \cref{sec: concept isolation}, respectively. Finally, in \cref{latent re-init} we discuss the implementation of latent re-initialization to achieve a more refined layout generation prior.
% Our emphasis lies on 
We emphasize integrating LoRAs to enable multi-concept customization within a single image, aiming for a solution that is both more flexible and scalable.

\subsection{LoRA-Composer Pipeline Overview}
\label{sec: pipeline overview}
As shown in \cref{fig:method}(a), LoRA-Composer utilizes a standard LoRA approach for subject registration, facilitating seamless integration of diverse subjects without requiring training for LoRA fusion. 
% Our approach to the intricate task of multi-subject customization unfolds in two steps: initially, we create an accurate representation of each subject, followed by their coherent combination into an image. The process begins by acquiring a LoRA model, which can be achieved through training or downloading one shared by the community. Following this, LoRA-Composer efficiently combines these acquired LoRAs into a unified, coherent image. 
Additionally, to further refine the model's capability in managing multiple conditions simultaneously, we provide the option to incorporate image-based conditions, using T2I-Adapter~\cite{mou2023t2i}. 

Our primary contribution is the introduction of the LoRA-Composer Block, depicted in \cref{fig:method}(b). 
In this innovation, we have re-designed the attention block within the U-Net architecture. Specifically, 
% we preserved the residual block structure while adapting both the self-attention and cross-attention mechanisms. 
in the cross-attention layers, we implement concept injection constraints designed to counteract concept vanishing.
Concurrently, within the self-attention layers, we introduce concept isolation constraints to effectively segregate different concepts, ensuring their distinctiveness. These strategies enable the refinement of the latent space into an image customized according to user preferences, utilizing both self-attention and cross-attention maps to direct the denoise process effectively.

%Starting with preparing the LoRA model, we train it using sets of images and corresponding text prompts for each subject, ranging from 5 to 15 pairs, to capture each subject's distinct features (or download from the community). Subsequently, LoRA-Composer merges these subjects into a single, cohesive image. Moreover, we offer the option to enhance our model's multi-condition control capabilities through the integration of T2I-Adapter~\cite{mou2023t2i}.

\subsection{Concept Injection Constraints}
\label{sec: concept injection}
% Simply using text prompts to specify desired concepts may result in missing concepts in the basic Stable Diffusion~\cite{chefer2023attend}. Although spatial attention guidance methods like BoxDiff~\cite{xie2023boxdiff}, Attend-and-Excite~\cite{chefer2023attend}, and Local control~\cite{zhao2023local} can mitigate the issue of missing objects in multi-concept generation, they fall short in precisely represent user-defined concepts. To tackle this challenge, we introduce the concept injection constraint, which is comprised of two key components: region-aware LoRA injection and concept enhancement constraints. 

% motivation: solve diminishing
Simply using text prompts to specify desired concepts may result in missing concepts in the basic Stable Diffusion~\cite{chefer2023attend}. Although spatial attention guidance methods like BoxDiff~\cite{xie2023boxdiff}, Attend-and-Excite~\cite{chefer2023attend}, and Local control~\cite{zhao2023local} can mitigate the issue of missing objects in multi-concept generation, they fall short in precisely represent user-defined concepts. Additionally, Mix-of-Show~\cite{gu2023mixofshow} involves optimizing a combination of lesser LoRA weights to preserve the characteristics of the concepts within the pre-trained model. However, this can diminish LoRA's capability to represent conceptual features, resulting in diminished concepts, as illustrated in \cref{fig:intro}. To tackle these challenges, we introduce the concept injection constraints, which is comprised of two key components: region-aware LoRA injection and concept enhancement constraints.

\begin{figure}[tb]
  \centering
  \includegraphics[width=\linewidth]{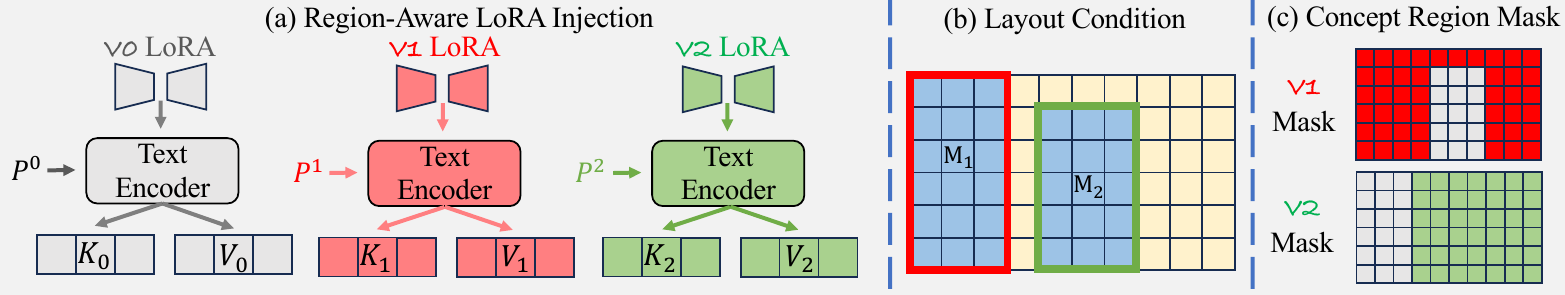}
    \vspace{-4mm}
  \caption{Modules of LoRA-Composer Block: (a) region-aware LoRA injection, (b) layout condition, (c) concept region mask, self-attention in the gray area is not calculated.
  }
  \label{fig: region_lora}
  \vspace{-5mm}
\end{figure}

\noindent\textbf{Region-Aware LoRA Injection.} 
% Drawing inspiration from regionally controllable sampling~\cite{gu2023mixofshow}, our method initiates with region-aware LoRA injection. 
Our approach directly injects each LoRA through region-aware LoRA injection, thereby avoiding the issue of missing concepts caused by the fusion of LoRAs.
As illustrated in \cref{fig: region_lora}(a), upon receiving a layout condition, as shown in \cref{fig: region_lora}(b), we extract the queries, keys, and values in the pre-defined layout $M_i$.
\begin{equation}
    Q_i = M_i \odot W_{0}^{Q}(z), K_i = W_{i}^{K}(\tau_{i}(P^{i})),  V_i = W_{i}^{V}(\tau_{i}(P^{i})),
\end{equation}
where $i \in \{0,1,...,N\}$. The pre-trained CLIP text encoder~\cite{radford2021learning} combined with LoRA is represented by $\tau_i$. As depicted in \cref{fig:method}(b), for $i \ne 0$, $i$ and $P^{i}$ correspond to the indices of foreground concept LoRAs and their associated local prompts, respectively. When $i=0$, these symbols refer to the background concept LoRA and the global prompt that describes the background. The symbol $\odot$ represents the Hadamard product. The $W^Q, W^K$, and $W^V$ stand for the projection matrices within the cross-attention modules of U-Net blocks which combined the concept LoRA.
After replicating this process for each concept, we then update the region's hidden state through the cross-attention mechanism as follows:
\begin{equation}
h_i = \text{softmax}\left( \frac{Q_i (K_{i})^{T}}{\sqrt{d}} \right) V_i,
\end{equation}
where the $d$ represents the dimension of queries and keys, this injection approach ensures a comprehensive integration of both background and foreground concepts, enhancing the model's ability to accurately reflect user-specified concepts within the generated images.

% \noindent \textbf{Concept Enhancement Constraints}: Building on the methodology presented by BoxDiff~\cite{xie2023boxdiff}, to ensure that synthesized objects accurately align with user-defined locations, a direct method involves confining high responses within the cross-attention mechanism to specified mask regions. Additionally, to address the issue of object generation tending towards the corners of the layout box, we incorporate a Gaussian weight into the cross-attention map sampling.

\noindent \textbf{Concept Enhancement Constraints.} Region-Aware LoRA Injection does not sufficiently stimulate all inherent abilities of each LoRA, consequently resulting in the loss of detailed characteristics (see \cref{fig: ablation}(e)). Previous methods~\cite{xie2023boxdiff, chefer2023attend} have proposed enhancing activation within the cross-attention mechanism in specific regions to improve the ability of Stable Diffusion~\cite{rombach2021highresolution} to represent concepts effectively.
However, these approaches are unsuitable for multi-concept customization tasks, as object generation tends toward the edges of the layout box (see Appendix \cref{fig: more_ablation}(a)) or fails to fill it adequately (see Appendix \cref{fig: more_ablation}(b)). To address these issues, we introduce a Gaussian weight into the cross-attention map, along with a $\mathcal{L}_{f}$ designed to ensure the fullness of elements within the box.

\begin{equation}
\label{equ:box_enhance}
    \mathcal{L}_{c} = \sum_{i=1}^{N}(1 - \frac{1}{S}  \sum_{j\in E} \textbf{topk}(A^{j}_i \odot M_i \odot G, S)),
\end{equation}
where $\textbf{topk}(\cdot, S)$ indicates the selection of $S$ elements with the highest response in input and $G$ means standard Gaussian distribution. We use the concept token by default ($V = [V_{rand}, V_{class}]$, for more details, see \cref{sec:Preliminary}) during the training of concept LoRA. $E$ refers to the indexes of concept token in prompts. For the foreground concept $i$, the cross-attention map corresponding to the $j$-th token is represented by $A^{j}_i$. 
% Moreover, we also propose the loss at the projection of the w-axis and h-axis, respectively, to ensure the concepts fill the entire box.
As for the $\mathcal{L}_{f}$, we squeeze cross-attention maps on the w-axis and h-axis via the max operation as below:
\begin{equation}
    a^{j}_{i}(w) = \mathbf{max}_{h=1,2,\dots,H}\{A^{j}_i(w,h)\},
\end{equation}
\begin{equation}
    a^{j}_{i}(h) = \mathbf{max}_{w=1,2,\dots,W}\{A^{j}_i(w,h)\},
\end{equation}
where the variables $W$ and $H$ denote the width and height of the
cross-attention map $A^{j}_i$. Then we compute the L1 loss in each axis as follows:
\begin{equation}
\label{equ:fill}
    \mathcal{L}_{f} = \frac{1}{L} \sum_{i=1}^{N}  \sum_{j\in E} (\mathbf{1} - \{M_i \odot a^{j}_{i}(w), M_i \odot a^{j}_{i}(h)\}),
\end{equation}
where $\{ \cdot, \cdot \}$ denotes the concatenation  followed by flattening and $\mathbf{1}$ represents a vector of ones. The term $L$ is defined as the length of $\mathbf{1}$.
% Overall, the implementation of the Inner-Box Concept Enhancement strategy is as follows:
% \begin{equation}
%     \mathcal{L}_{e} = \alpha \mathcal{L}_{c} + \beta \mathcal{L}_{f}
% \end{equation}

\subsection{Concept Isolation Constraints}
\label{sec: concept isolation}
While the concept injection constraints effectively guarantee that objects will be placed within user-specified regions, they cannot prevent the potential overlap or infection of customized concepts within these areas (see \cref{fig: ablation}(d)). To preserve the integrity and distinctiveness of each concept within its designated region, we introduce concept isolation constraints. This approach is divided into two main components: concept region mask and region perceptual restriction. Both elements are integrated within the self-attention mechanism of the U-Net block, ensuring that each concept remains isolated and unaffected by others, thereby maintaining the purity of concepts in their target regions.

\noindent \textbf{Concept Region Mask.} The self-attention mechanism creates connections among all query elements, essential for maintaining the distinct characteristics of each concept. To preserve the distinctiveness of each concept, we adopt the concept region mask strategy, guided by a given layout condition as depicted in \cref{fig: region_lora}(b). This design limits the interaction between queries within a specific concept region and those in other concept regions, as demonstrated in \cref{fig: region_lora}(c). Consequently, it ensures the preservation of each concept's characteristics, free from the influence of neighboring concepts.

\noindent \textbf{Region Perceptual Restriction.} Due to down-sampling and residual convolution operations in the U-Net framework, concept features might spread into the elements designated for background areas, as highlighted by the yellow square in \cref{fig: region_lora}(b). To mitigate the risk of concept feature leakage into unintended regions, we employ region perceptual restriction, aimed at minimizing interaction between queries of the foreground and background areas. This technique ensures that each concept remains distinct and unaffected by the features of the background feature, thereby preserving the uniqueness and integrity of each concept within the synthesized image. This formulated as
\begin{equation}
\label{equ: region}
    \mathcal{L}_{r} = \frac{1}{S} \sum_{i=1}^{N} \textbf{topk}(\Bar{A}{[M_i,\mathbf{1}-{M}_i]}, S),
\end{equation}
where the $\Bar{A}{[M_i,\mathbf{1}-{M}_i]}$ refers to the self-attention map obtained through a matrix slicing operation across the channel dimension.
% and the $\Tilde{M}_n  \in \mathbb{R}^{(\Tilde{B}, \Tilde{C})}$ represent the NOT operation on $M_n$. Specifically, the original self-attention map $\hat{A} \in \mathbb{R}^{(L*S, L*S)}$ undergoes a slicing operation to yield $\Bar{A}_{[M_n,\Tilde{M}_n]}\in \mathbb{R}^{(B*C, \Tilde{B}*\Tilde{C})}$, where the $L*S$ denotes the length of the query vector. 

At each timestep, overall constraints loss are formulated as:
\begin{equation}
    \mathcal{L} = \mathcal{L}_{c} + \alpha \mathcal{L}_{f} + \beta \mathcal{L}_{r},
\end{equation}
where $\alpha$ and $\beta$ represent weighting coefficients. Using the constraints loss $\mathcal{L}$, the current latent $z_t$ can be updated with a step size of $\phi_{t}$ as follow:
\begin{equation}
\label{equ:loss}
    z'_{t} \leftarrow z_{t} - \phi_{t} \cdot \nabla \mathcal{L}.
\end{equation}
Following BoxDiff~\cite{xie2023boxdiff}, the step size $\phi_{t}$ decays linearly with each timestep. By incorporating the previously mentioned constraints, $z'_t$ is directed at each timestep to foster the generation of customized concepts within designated locations, while preventing the leakage of concept features into areas associated with other concepts. Subsequently, $z'_t$ is utilized as the input for the U-Net for the ensuing inference step $z'_{t} \xrightarrow{U-Net} z_{t-1}$. This strategic guidance ensures the precise synthesis of target concepts within the user-specified layout regions.

\subsection{Latent Re-initialization}
\label{latent re-init}
We discovered that traditional LoRA is not ideally suited for generating specific local areas, because it is trained without control in location. This discrepancy can result in imprecise locations for concept generation (see \cref{fig: ablation}(c)). To address this issue, we propose re-initializing the latent space to better accommodate the integration of concept-specific LoRAs. 

% Our approach aims to identify the position within the latent space where the object is likely to appear and then align this position with the specified layout. Specifically, before the denoising phase, we initialize the latent space $z_t$ with Gaussian noise and apply the LoRA-Composer process for a one-step update using \cref{equ:loss}. Afterward, a cross-attention map is generated. We then replace the layout area $z_{t}{[M_i]} = z_{t}[\hat{x}_{i}, \hat{y}_{i}] $ with the highest scoring area. The top-left coordinate of area is determined as follows:

% \begin{equation}
%     \hat{x}_{i}, \hat{y}_{i} = \underset{x \in 0,1 \dots w, y \in 0,1 \dots h} {\mathrm{arg\,max}}\ \sum_{j=0}^{\mathbb{W}} \sum_{k=0}^{\mathbb{H}} [\textbf{crop}(A_{i},[x,y],\mathbb{W},\mathbb{H})]_{[j,k]},
% \end{equation}
% where $\mathbb{W}$ and $\mathbb{H}$  represent the width and height of the concept mask $M_i$, respectively. The variables $w$ and $h$ denote the width and height of the cross-attention map $A_{i}$. The function $\textbf{crop}(\cdot ,[x,y],\mathbb{W},\mathbb{H})$ refers to cropping the attention map to a shape of $\mathbb{W},\mathbb{H}$ with the top-left coordinate at $[x,y]$. The notation $[\cdot]_{j,k}$ denotes the element located at row $j$ and column $k$ in the matrix. Finally, the latent is normalized to a standard Gaussian distribution to prepare for subsequent processing steps.

Our approach aims to identify the position within the latent space where the object is likely to appear and then align this position with the specified layout. Specifically, before the denoising phase, we initialize the latent space $z_t$ with Gaussian noise and apply the LoRA-Composer process for a one-step update using \cref{equ:loss}. Afterward, a cross-attention map is generated based on the region latent query $Q_i$ and the textual embedding $K_i$ for each local prompt. We first compute each candidate area with the same shape as the layout. We then replace the layout area $z_{t}{[M_i]}$ with the highest scoring area in the candidate areas. Finally, the latent is normalized to a standard Gaussian distribution to prepare for subsequent processing steps. The aforementioned candidate areas can be expressed as:
\begin{equation}
\label{equ: re_init}
    \hat{A_{i}} = \{\Phi(A_{i},[x,y],\mathbb{W},\mathbb{H})\}, 
\end{equation}
where $x \in \{0,1 \dots w\}, y \in \{0,1 \dots h\}$. The $w$ and $h$ denote the width and height of the cross-attention map $A_{i}$. The function $\Phi (\cdot,[x,y],\mathbb{W},\mathbb{H})$ refers to cropping the attention map to a shape of $\mathbb{W},\mathbb{H}$ with the top-left coordinate at $[x,y]$. The variables $\mathbb{W}$ and $\mathbb{H}$ represent the width and height of the layout box.

\begin{figure}[!t]
  \centering
  \includegraphics[width=\linewidth]{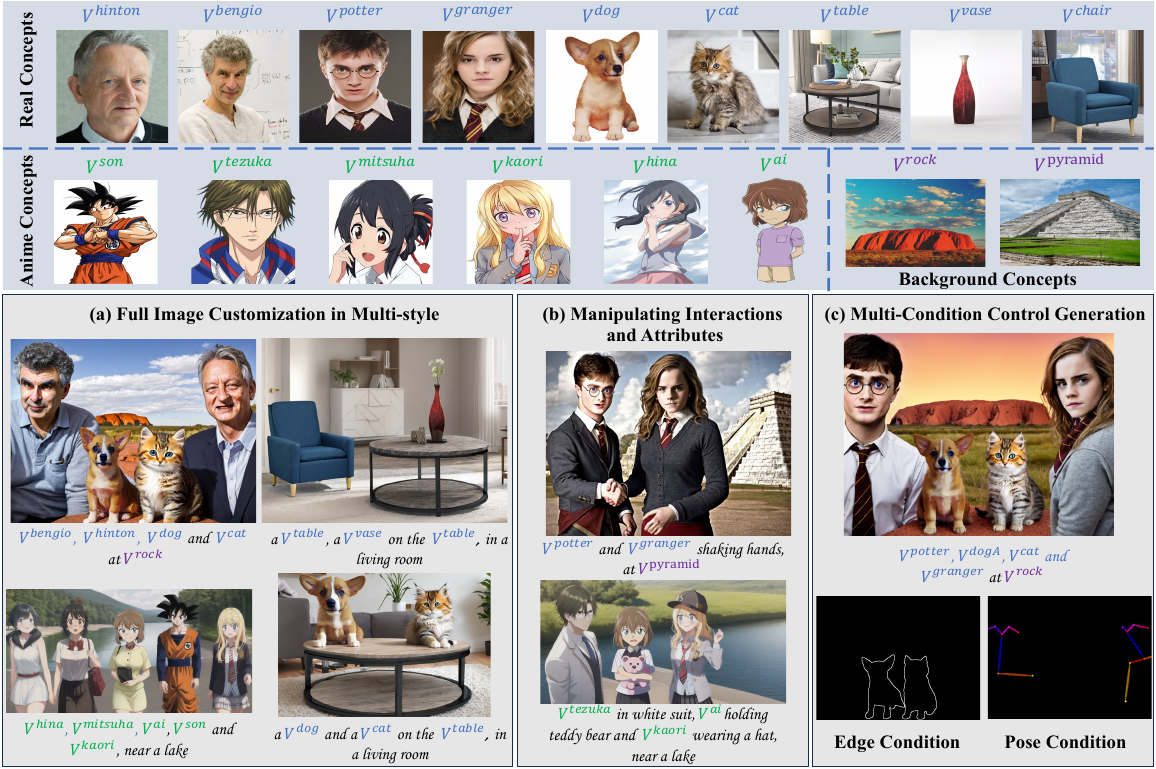}
  \vspace{-3mm}
  \caption{Three highlights of LoRA-Composer, a) full image customization in multi-style; b) manipulating interactions and attributes; c) multi-condition control generation.
  }
  \label{fig: main_res}
  \vspace{-4mm}
\end{figure}

\section{Experiments}

\subsection{Experimental Setup}
\noindent\textbf{Datasets}. For a thorough evaluation of LoRA-Composer, follow Mix-of-Show~\cite{gu2023mixofshow}, we compile a dataset featuring characters, animals, and backgrounds in both realistic and anime styles, encompassing a total of 16 customized subjects (see Appendix \cref{fig: dataset}). Through comprehensive experimentation across diverse subject combinations, we demonstrate the superior performance of our approach.

\noindent \textbf{Evaluation metrics}. Our evaluation of the LoRA-Composer approach utilizes two metrics for customized generation, as proposed in prior studies~\cite{cones2,gu2023mixofshow,kumari2022multiconcept}. (1) Image similarity, which assesses the visual resemblance between the generated images and the target subjects in the CLIP~\cite{radford2021learning} image embedding. 
% In our multi-concept generation task, we compute the image similarity for the generated images about each target concept individually and then determine the average value. 
(2) Textual similarity, which evaluates the similarity between generated images and their associated textual prompts, utilizing the CLIP model~\cite{radford2021learning}. 
% For our multi-concept generation task, for the foreground concepts, we first isolate each concept within the generated images. Subsequently, we extract every concept token $V$ in local prompt $P^{i}$ and substitute it with the concept's class name before computing the textual similarity. When evaluating background concepts, we input the entire image. The final score is the average of these similarity values.
Please refer to \cref{Implementation} for details.

\noindent \textbf{Baseline}. 
% To assess the quality of our generated images, 
We compare our approach against four leading competitors in the field. \textbf{Cones2}~\cite{cones2} leverages text embeddings to support arbitrary combinations of concepts.
% without necessitating model refinement. 
\textbf{Mix-of-Show}~\cite{gu2023mixofshow} employs gradient fusion to integrate multiple concepts into a base model. Both \textbf{Anydoor}~\cite{chen2023anydoor} and \textbf{Paint by Example}~\cite{yang2023paint} facilitate multi-concept generation by utilizing networks trained specifically for inpainting tasks. 
More implementation details are shown in the \cref{Implementation}.

\subsection{Visualization Results}
% Our methodology enables extensive customization of the entire image, addressing both foreground and background components and accommodating a wide range of styles, from anime to realistic. 
Our broad customization capability is illustrated in \cref{fig: main_res}(a), showcasing our approach's versatility in adapting to a wide range of styles, from anime to realistic. 
% adapting to artistic expressions. 
Additionally, our method enables precise manipulation of interactions and attributes, such as shaking hands, wearing hats, and holding teddy bears in the picture, directly through textual prompts. This capability is showcased in \cref{fig: main_res}(b). Moreover, our framework is designed for flexibility, capable of generating images under multiple conditions. It adeptly integrates specific constraints such as edge detection or pose estimation to guide the image synthesis process. This capacity for accommodating additional image-based conditions, as detailed in \cref{fig: main_res}(c), highlights the adaptability of our approach in meeting varied and complex generation requirements. More visualization results are shown in \cref{fig: more_res} in the Appendix.

\begin{table}[!t]
    \centering
    \scriptsize
    \resizebox{0.8\linewidth}{!}{
    \begin{tabular}{lcccccc}
    \toprule
        \textbf{Method}
        & \textbf{Anime-I}
        & \textbf{Anime-T}
        & \textbf{Real-I} 
        & \textbf{Real-T}
        & \textbf{Mean-I}
        & \textbf{Mean-T} \\
\midrule 
 Cones2~\cite{cones2} & 0.5940 & 0.5691 & 0.5924 & 0.5912 & 0.5883 & 0.5807 \\
 Mix-of-Show~\cite{gu2023mixofshow} & 0.6296 & 0.5741 & 0.6743 & 0.5970 & 0.6519 & 0.5856 \\
 Anydoor~\cite{chen2023anydoor} & - & - & 0.6801 & \textbf{0.6410} & 0.6801 & \textbf{0.6410} \\
 Paint by Example~\cite{yang2023paint} & - & - & 0.6888 & 0.6356 & 0.6888 & 0.6356 \\
 \textbf{LoRA-Composer} & \textbf{0.8219} & \textbf{0.5945} & \textbf{0.7399} & 0.6278 & \textbf{0.7809} & 0.6111 \\
\midrule 
 Mix-of-Show*~\cite{gu2023mixofshow} & 0.8238 & \textbf{0.6067} & 0.7097 & 0.6203 & 0.7668 & 0.6135 \\
 \textbf{LoRA-Composer*} & \textbf{0.8320} & 0.5981 & \textbf{0.7367} & \textbf{0.6314} & \textbf{0.7843} & \textbf{0.6147} \\
\bottomrule
\end{tabular}}
\vspace{2mm}
\caption{Quantitative comparison of LoRA-Composer with baselines in generating anime and realistic style concepts. \textbf{T} refers to textual similarity, \textbf{I} refers to image similarity, with an asterisk \textbf{*} indicating the use of image-based conditions. The highest scores in each column are marked in \textbf{bold} for clarity.}
\label{tab:eval_simple}
\vspace{-3mm}
\end{table}

\subsection{Quantitative Results}
As detailed in \cref{tab:eval_simple}, our LoRA-Composer surpasses prior methods in image similarity, showcasing its effectiveness across both anime and realistic styles. Conversely, inpainting-based methods such as Anydoor and Paint by Example exhibit higher text similarity. This is because these methods specialize in inserting subjects into user-defined locations through reference images, focusing more on aligning with textual descriptions. Our method achieves significant improvements over Cones2, thanks to the efficacy of our LoRA-Composer block.
% These data illustrate that our method excels in precisely drawing out target concepts.

To ensure fairness in our comparison, we established two settings: one without using image-based conditions and another incorporating them (indicated by *). Our method consistently outperforms in both scenarios. We observed that image-based conditions play a crucial role in Mix-of-Show. Without these conditions, it faces severe drops in both image and textual similarity. In contrast, LoRA-Composer exhibits enhanced robustness and gains further advantage from image-based conditions, offering increased convenience to users.

\begin{figure}[!t]
  \centering
  \includegraphics[width=0.85\textwidth]{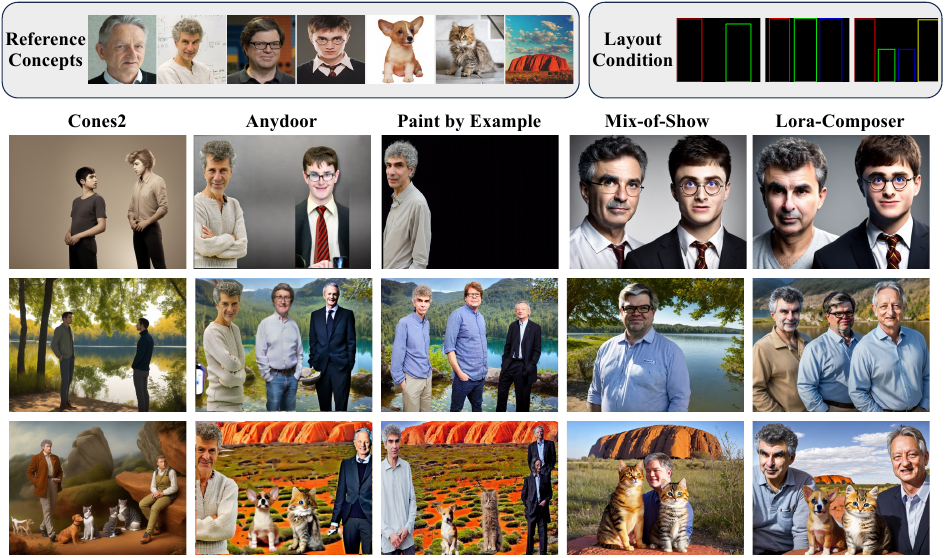}
  \vspace{-2mm}
  \caption{Qualitative comparison with baselines. For each case, we use the same seeds.}
  \vspace{-5mm}
  \label{fig: comparison}
\end{figure}

\vspace{-3mm}
\subsection{Qualitative Comparison}
LoRA-Composer is evaluated without using image-based conditions across four benchmarks in multi-concept customization scenarios. The results are displayed in \cref{fig: comparison}. It can be seen that Cones2~\cite{cones2} faces concept confusion, unable to clearly capture concept characteristics. In contrast, inpainting-based models like Anydoor~\cite{chen2023anydoor} and Paint by Example~\cite{yang2023paint} retain certain concept features, such as glasses and gray curly hair as illustrated in the first row, but struggle with detailed attributes, especially facial features. 
These approaches also lead to disproportionate foreground elements and unnatural integration of foreground and background. 
Mix-of-Show~\cite{gu2023mixofshow} also suffers from concept confusion, as seen in the first row where glasses are incorrectly placed on a man to the left.
Furthermore, in the second and third rows, two people are completely vanishing. 
Differently, our method successfully synthesizes images that accurately incorporate all subjects with their correct characteristics, showcasing enhanced performance in multi-concept synthesis and attribute accuracy.
Additional qualitative comparison results and results using image-based condition guidance are provided \cref{fig: more_comparison} and \cref{fig: condition_comparison} in the Appendix. 

\subsection{Ablation Study}
\label{sec: ablation study}
\begin{figure}[!t]
  \centering
  \includegraphics[width=\textwidth]{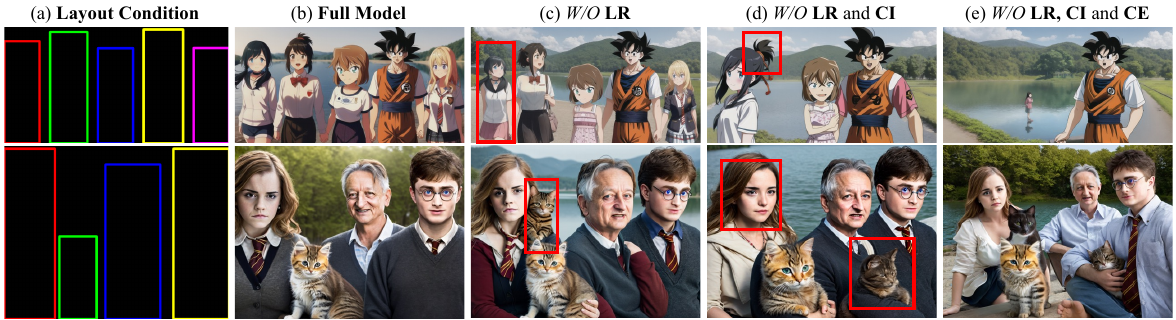}
  \vspace{-4mm}
  \caption{Visualized results from ablation study on individual components. \textbf{``LR''} stands for latent re-initialization, \textbf{``CI''} denotes concept isolation constraints, and \textbf{``CE''} signifies concept enhancement constraints within concept injection constraints.
  }
  \vspace{-6mm}
  \label{fig: ablation}
\end{figure}

To demonstrate the efficacy of each component in our method, we choose a challenging scenario with five and four concepts for this section. As illustrated in \cref{fig: ablation}(c), the positions of the anime girl and the cat within the red box diverge from the layout conditions outlined in \cref{fig: ablation}(a). This discrepancy serves as evidence that omitting latent re-initialization (LR) hampers the precise placement of concepts, mainly because of the lack of spatial priors. Subsequently, as shown in \cref{fig: ablation}(d), removing the concept isolation constraints (CI) results in the blending of concept characteristics. This results in observable issues such as the confusion in the anime girl's haircut, the distortion of the woman’s face, and the appearance of another cat in an incorrect area. Without CI, concepts begin to overlap and influence each other, resulting in a disruption of harmony and coherence in the overall image composition. Finally, as shown in \cref{fig: ablation}(e), eliminating the concept enhancement constraints (CE) results in the disappearance of concepts. However, thanks to the presence of region-aware LoRA injection, the model retains the capability to insert concepts, though with diminished precision in their placement and representation. This highlights each element's critical role in achieving precise and harmonious concept integration.

% 之前做错的版本
% To substantiate our findings as non-coincidental, 
% we conducted a quantitative evaluation, as presented in \cref{tab: ablation}. 
% latent re-initialization (LR) leads to notable enhancements across all performance metrics, owing to its effectiveness in refining region-specific priors. Similarly, the concept enhancement constraints (CE) significantly improve all metrics due to their capacity to activate concepts more effectively. Moreover, concept isolation constraints (CI) also positively influence the metrics, underscoring their importance in boosting the model's overall performance. The synergistic integration of these three components culminates in optimal outcomes.
To substantiate our findings as non-coincidental, we conducted a comprehensive quantitative evaluation. As detailed in \cref{tab: ablation}, our analysis demonstrates the individual and collective impacts of concept enhancement constraints (CE), latent re-initialization (LR), and concept isolation constraints (CI) on performance. CE lead to significant improvements across all performance metrics, showcasing its effectiveness in activating concepts, evidenced by an increase of around 0.1 in mean image similarity. LR further contributed to these enhancements by refining region-specific priors. CI played a crucial role in preserving the distinctiveness of concept traits and enhancing model robustness. More ablation results are shown in \cref{More Ablation Study} in the Appendix.

\begin{table}[!t]
    \centering
    \scriptsize
    \resizebox{0.85\linewidth}{!}{
    \begin{tabular}{ccccccccc}
    \toprule
        \textbf{CE}
        & \textbf{CI}
        & \textbf{LR}
        & \textbf{Anime-I}
        & \textbf{Anime-T}
        & \textbf{Real-I} 
        & \textbf{Real-T}
        & \textbf{Mean-I}
        & \textbf{Mean-T} \\
\midrule 
 \rowcolor{gray!40} \checkmark & \checkmark & \checkmark & 0.8031 & 0.5948 & 0.7347 & 0.6319 & 0.7739 & 0.6134 \\
 \checkmark & \checkmark & & 0.8024 & 0.5923 & 0.7350 & 0.6314 & 0.7687 & 0.6119 \\
 \checkmark &  &  & 0.7957 & 0.5899 & 0.7271 & 0.6250 & 0.7614 & 0.6075 \\
 &  &  & 0.6597 & 0.5725 & 0.6683 & 0.6105 &0.6640 & 0.5915 \\
\bottomrule
\end{tabular}}
\vspace{2mm}
\caption{Ablation studies on various components. \textbf{"LR"} stands for latent re-initialization, \textbf{"CI"} denotes concept isolation constraints, and \textbf{"CE"} signifies concept enhancement constraints within concept injection constraints.}
\label{tab: ablation}
\vspace{-4mm}
\end{table}

\section{Conclusion}
In this paper, we introduce LoRA-Composer, a novel approach designed to seamlessly integrate multiple concepts within a single image. We explore two prevalent issues in multi-concept customization: concept vanishing and concept confusion. 
% To mitigate these challenges, 
To this end, we employ concept injection constraints to combat concept vanishing, while concept isolation constraints alleviate concept confusion. Additionally, we propose latent re-initialization to provide precise region priors. Our experiments highlight the capability of LoRA-Composer to customize entire images, including both background and foreground elements, and to manipulate the interactions and attributes of various concepts through textual prompts. Compared to traditional methods, LoRA-Composer offers enhanced flexibility and usability, allowing users to generate images with fewer conditions and readily accessible LoRA techniques. Furthermore, we demonstrate the method's ability to achieve high-fidelity combinations of multiple concepts, underscoring its practical utility in complex image-generation tasks.

\bibliographystyle{unsrt}
\bibliography{main}

\clearpage

%%%%%%%%%%%%%%%%%%%%%%%%%%%%%%%%%%%%%%%%%%%%%%%%%%%%%%%%%%%%

\appendix

Considering the space limitation of the main text, we provided more results and discussion in this supplementary material, which is organized as follows:
\begin{itemize}
    \item Section {\ref{sec:Preliminary}}: a concise overview of diffusion models~\cite{rombach2021highresolution} and ED-LoRA~\cite{gu2023mixofshow}.
    \item Section {\ref{Implementation}}: implementation details of our approach and the baseline models.
    \item Section {\ref{Experiments}}: more detailed experiments analysis and discussion.
        \begin{itemize}
          \item Section {\ref{default setting}}: our default setting in experiments and ablation study.
          \item Section {\ref{More Ablation Study}}: ablation study on concept enhancement constraints (in \cref{sec: concept injection}) and concept isolation constraints (in \cref{sec: concept isolation}).
          \item Section {\ref{More Comparison}}: comparison with Mix-of-Show, under the image-based conditions.
          \item Section {\ref{User Study}}: assess the human preference between our method and baseline approaches.
          \item Section {\ref{More Visual Results}}: more visual results of LoRA-Composer.
        \end{itemize}
    \item Section {\ref{Potential Negative Society Impact}}: discussion of our method's potential negative society impact.
    \item Section {\ref{Limitation}}: failure cases and discussion.

\end{itemize}

\section{Preliminary}
\label{sec:Preliminary}
\textbf{Diffusion Models} are famous for their capacity to generate high-quality images. Their framework operates in two primary phases: the forward phase, where Gaussian noise is progressively added to an image until it fully conforms to a Gaussian distribution, and the reverse phase, which aims to reconstruct the original image from its noised condition. The reverse phase typically employs a U-Net architecture enhanced with text conditioning, enabling the synthesis of images based on textual descriptions during inference. In this work, we employ Stable Diffusion~\cite{rombach2021highresolution}, which distinguishes itself by operating in the latent space rather than directly manipulating image pixels through these phases. This approach involves an autoencoder, with an encoder $\mathcal{E}$ and decoder $D$, trained to serve as a bridge between image pixel space $x$ and latent space $z$, i.e., $D(z) = D(\mathcal{E}(x))$. In each time step $t$, given a textual condition $\tau(P)$ and an image $x$, where 
$P$ represents the text prompt and $\tau$ denotes the pre-trained CLIP text encoder~\cite{radford2021learning}. The training objective for Stable Diffusion is to minimize the denoising objective by

\begin{equation}
    \mathcal{L}_{sd} = \mathbb{E}_{z \sim \mathcal{E}(x), P, \epsilon \sim \mathcal{N}(0,1), t} \left[ \| \epsilon - \epsilon_{\theta}(z_t, t, \tau(P)) \|_2^2 \right],
\end{equation}
where $\epsilon_{\theta}$ is the denoising U-Net with learnable parameter $\theta$.

\noindent\textbf{ED-LoRA}~\cite{gu2023mixofshow}
aims to augment the expressiveness of the embedding by employing a decomposed structure. ED-LoRA implements a layer-wise embedding strategy, following the method described in P+~\cite{voynov2023p+}, to forge a multi-faceted representation for the concept token ($V=[V_{rand}, V_{class}]$). This involves adding a random variation ($V_{rand}$) and a class-specific component to the base embedding ($V_{class}$). Furthermore, it integrates a LoRA layer into the linear layers present in all attention modules of the text encoder and U-Net. This integration allows for a flexible adaptation of the model to specific concepts by modifying the linear layers in a low-rank manner, thereby enhancing the model's ability to encode and synthesize images based on textual descriptions with high fidelity. We use it by default in all experiments.

\section{Implementation Detail}
\label{Implementation}

\noindent\textbf{ED-LoRA Setting.} We chose ED-LoRA due to its strong capability in maintaining concept fidelity. In alignment with the single-concept ED-LoRA tuning guidelines from~\cite{gu2023mixofshow}, we integrate the LoRA layer into the linear layers of all attention modules within both the text encoder and U-Net, setting a rank (\(r\)) of 4 for all experiments. For optimization, we employ the Adam optimizer~\cite{Kingma2014AdamAM}, utilizing learning rates of 1e-5 for the text encoder and 1e-4 for U-Net tuning.

\noindent\textbf{Sample Details.} For all experiments and evaluations in this paper, we use the DPM-Solver~\cite{DPM-Solver}, implementing adaptive sampling steps to enhance computational efficiency. Specifically, if the loss (as described in \cref{equ:loss}) ceases to decrease, we stop the process, thereby accelerating the overall procedure. For this loss, the relative coefficients are set as $\alpha = 0.25$ and $\beta = 0.8$. 

% \subsection{Evaluation Setting} % prompt setting \lora from mos and our
% \label{Evaluation Setting}
\begin{figure}[!tb]
  \centering
  \includegraphics[width=\textwidth]{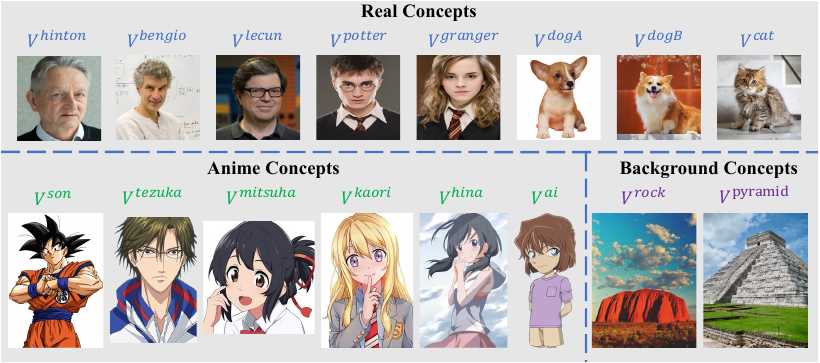}
  \vspace{-5mm}
  \caption{The datasets utilized for our model encompass a diverse range of concepts, including real-world objects, anime characters, and background scenes, totaling 16 distinct concepts.}
  \vspace{-5mm}
  \label{fig: dataset}
\end{figure}

\textbf{Pretrained Models.} Following the approach used by Mix-of-Show~\cite{gu2023mixofshow}, we select Chilloutmix\footnote{https://huggingface.co/windwhinny/chilloutmix} as the pre-trained model for crafting real-world concept images. For anime concepts, Anything-v4\footnote{https://huggingface.co/xyn-ai/anything-v4.0} serves as the chosen pre-trained model. To ensure equitable comparisons among different methods, all comparative analyses involving training-based methods, such as Cones2~\cite{cones2} and Mix-of-Show~\cite{gu2023mixofshow}, utilize these specified pre-trained models, guaranteeing uniformity in evaluation criteria. For inpainting-based models, specifically Anydoor~\cite{chen2023anydoor} (which refines Stable Diffusion v2.1) and Paint by Example~\cite{yang2023paint} (which refines Stable Diffusion 1.4), we adhere to their official models.

\noindent\textbf{Evaluation metrics.} In our multi-concept generation task, we crop each foreground concept within the generated images, while using the entire image for the background. We compute the image similarity for each target concept individually and then calculate the average value. In addition, we extract every concept token $V$ in local prompt $P^{i}$ and substitute it with the concept's class name before computing the textual similarity. The final score is the average of these similarity values.

\noindent\textbf{Baseline Implementation Detail.} For \textbf{Cones2}~\cite{cones2}, we utilize the official implementation provided at the repository\footnote{https://github.com/ali-vilab/Cones-V2}. The training configurations specified include a batch size of 4, a learning rate of 5e-6, and a total of 4000 training steps. This setup requires approximately 10-15 minutes to execute on a single NVIDIA A100 GPU. To ensure consistency across experiments, we employ the same seed for image generation. Given that \textbf{Paint by Example}~\cite{yang2023paint}\footnote{https://github.com/Fantasy-Studio/Paint-by-Example} and \textbf{Anydoor}~\cite{chen2023anydoor}\footnote{https://github.com/ali-vilab/AnyDoor} focus exclusively on real object inpainting, we ensure a fair comparison by limiting the comparison to real-world concepts. Specifically, our approach involves initially generating the background image using our model with the same prompt and seeds, while omitting the foreground prompts. Subsequently, their models are employed to introduce the foreground concepts. For \textbf{Mix-of-Show}~\cite{gu2023mixofshow}, we utilize the same LoRAs for both real-world and anime concepts. We apply their gradient fusion technique\footnote{https://github.com/TencentARC/Mix-of-Show/tree/research\_branch} to integrate all of the LoRAs into the base model. Consistency across experiments is ensured by using the same seed for image generation, allowing for a direct comparison of outcomes.

\section{Additional Experiments}
\label{Experiments}

\subsection{Default setting in experiments}
\label{default setting}
We collect a diverse dataset featuring characters, animals, and backgrounds in both realistic and anime styles, encompassing 16 unique subjects (as shown in \cref{fig: dataset}). To assess our model, we randomly picked three varied settings in two styles, testing combinations of two to five subjects. We produced 50 images for each setting, culminating in $2 \times 3 \times 4 \times 50 = 1200$ images for an extensive performance review.    

For our ablation study, we selected three challenge settings within both anime and realistic styles, involving four and five concepts. This approach yielded 600 images, offering a substantial dataset to examine the effects of different model components and settings on our framework's capability.

\subsection{More Ablation Study}
\label{More Ablation Study}

\begin{figure}[!tb]
  \centering
  \includegraphics[width=\textwidth]{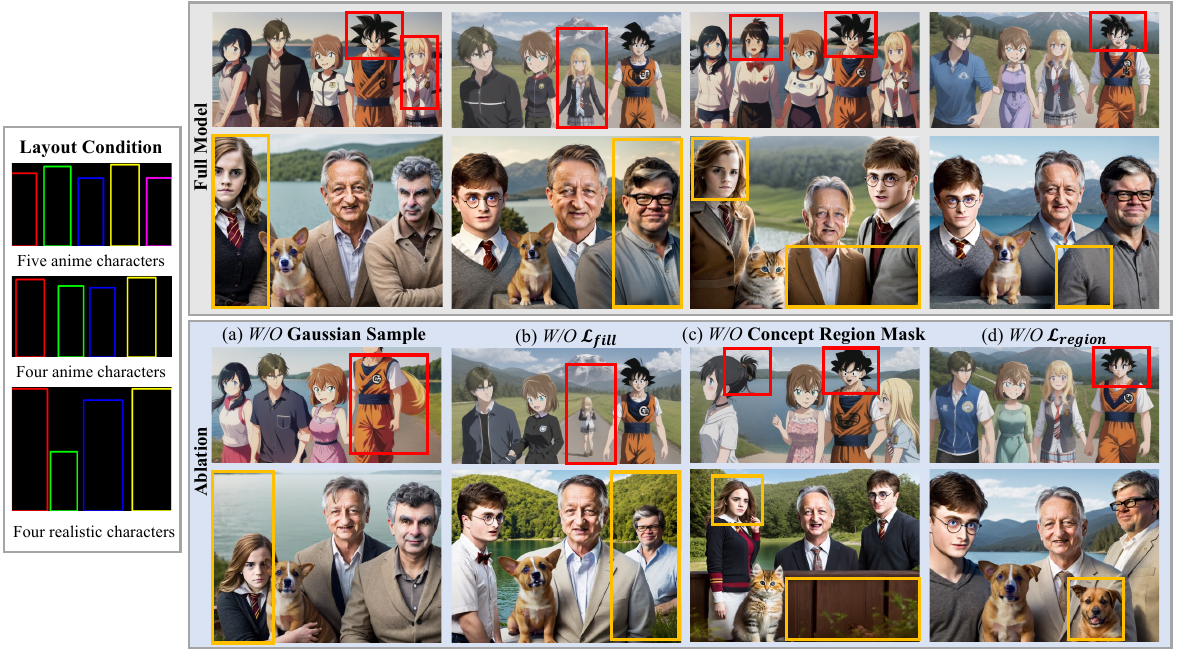}
  \vspace{-5mm}
  \caption{More ablation study on concept enhancement constraints and concept isolation constraints. The upper portion displays outcomes utilizing our full methodology, while the lower portion illustrates results with specific modules omitted, highlighting the significance of each component within our approach. The red boxes and the yellow boxes are used to accentuate the distinctions between the anime style and the real-world style, respectively.}
  \vspace{-5mm}
  \label{fig: more_ablation}
\end{figure}

In the main ablation study (\cref{sec: ablation study}), we explored the synergistic effects of combining modules with similar functionalities. Here, we delve deeper with an extensive ablation study on our concept enhancement constraints, which includes Gaussian sample strategy in \cref{equ:box_enhance} and $\mathcal{L}_{f}$ in \cref{equ:fill}) and concept isolation constraints (incorporating the concept region mask and $\mathcal{L}_{r}$ in \cref{equ: region}). These examinations aim to illuminate their contributions to model performance, as illustrated in \cref{fig: more_ablation}. Specifically, in the first column, the absence of Gaussian sampling leads to the concepts not being accurately centered within their designated boxes. This lack of precision can even cause anime concepts to appear outside their intended boundaries, resulting in a loss of their unique identity traits. In the second column, without $\mathcal{L}_{f}$, both anime and realistic figures fail to occupy their designated boxes completely, pointing to a deficiency in fully utilizing the allocated space. In the third column, we observe concept confusion, characterized by the merging of anime haircuts and the loss of distinctive facial traits in realistic figures, which indicates a loss of distinctiveness. This highlights the role of the concept region mask in safeguarding each concept's unique attributes. In the last column, concept features, such as the anime boy’s haircut being influenced by another character, and an unintended dog appearing. These issues indicate that there is leakage into unintended areas, due to down-sampling in U-Net. The inclusion of $\mathcal{L}_{r}$ effectively addresses this problem by minimizing the influence of background elements on foreground concepts. These strategies validate the essential roles played by the concept enhancement and concept isolation constraints in maintaining concept integrity and precision within the generated images, significantly bolstering the model's capability to produce conceptually coherent and visually accurate outputs.

\subsection{More Comparison}
\label{More Comparison}

\begin{figure}[!tb]
  \centering
  \includegraphics[width=\textwidth]{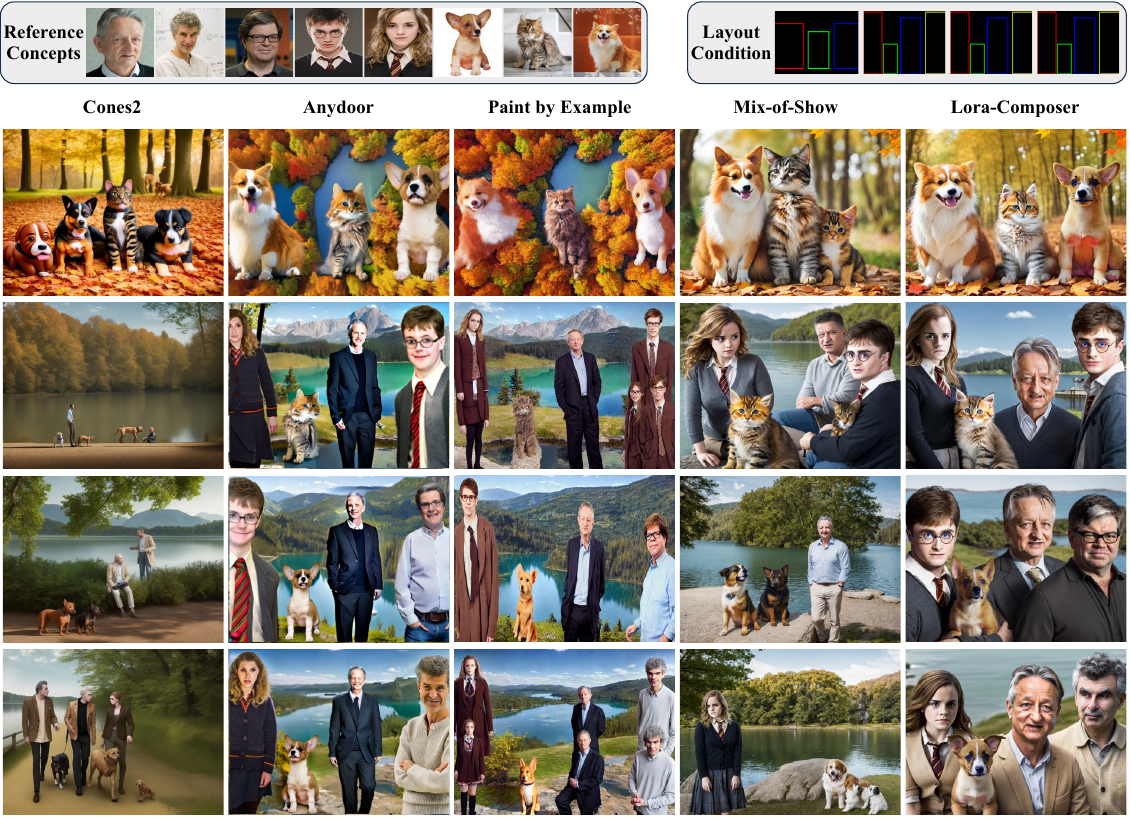}
  \caption{More qualitative comparison with baselines. For each case, we use the same seeds.}
    % \vspace{-3mm}
  \label{fig: more_comparison}
\end{figure}

\begin{figure}[!tb]
  \centering
  \includegraphics[width=\textwidth]{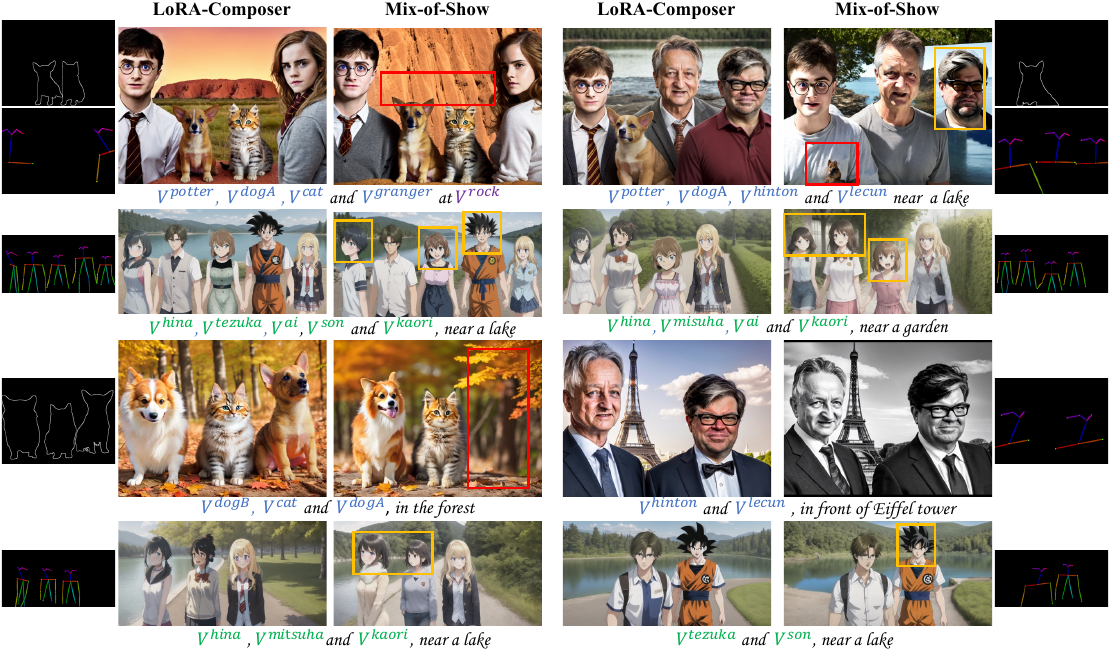}
  \caption{Comparison with Mix-of-Show, where image-based conditions are applied. The yellow box emphasizes the issue of concept confusion, while the red boxes underscore instances of concept vanishing.}
    % \vspace{-3mm}
  \label{fig: condition_comparison}
\end{figure}

To ensure a fair comparison with Mix-of-Show~\cite{gu2023mixofshow}, we adopted their default settings, applying the same image-based conditions and using identical random seeds for generating multi-concept images. The comparative results are depicted in \cref{fig: condition_comparison}, showcasing four unique concept combinations styled in both anime and real-world visuals. Our analysis reveals that while Mix-of-Show struggles with maintaining distinct identity features (as indicated by yellow boxes) and the completeness of the integrated concepts (highlighted by red boxes), our approach successfully overcomes these limitations. Our method produces high-fidelity, coherent images that significantly enhance user satisfaction and improve the perceived quality of the generated content.

\begin{figure}[!tb]
  \centering
  \includegraphics[width=0.8\textwidth]{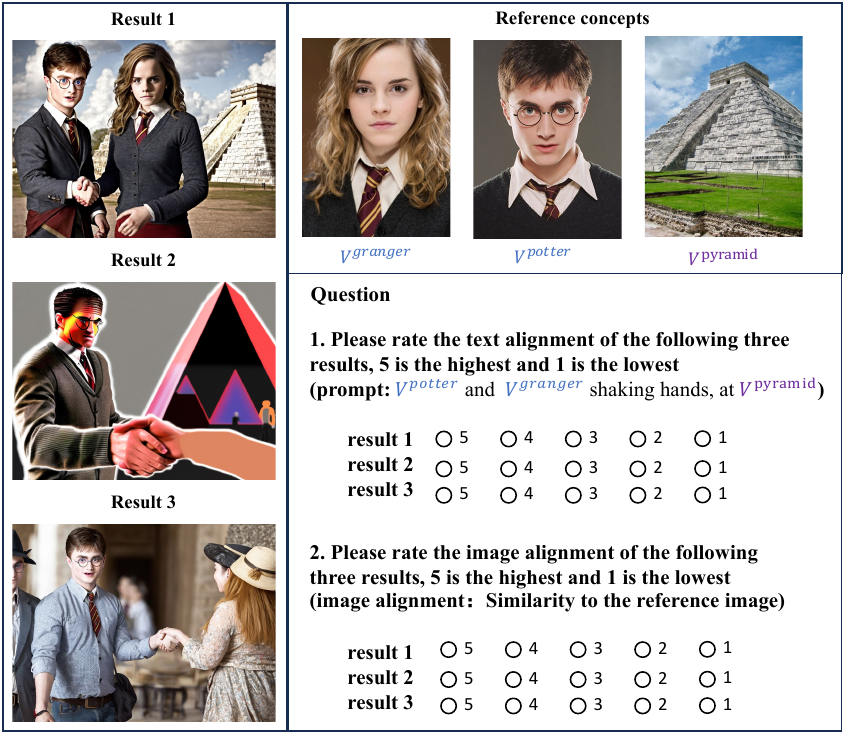}
  \caption{The user study interface that participants used to evaluate generated images on text-to-image alignment and image-to-image alignment.}
    \vspace{-2mm}
  \label{fig: user_study}
\end{figure}

\subsection{User Study}
\label{User Study}

To assess the efficacy of our multi-object customization outcomes more accurately, we implemented a user study to capture human preferences. Following the approach utilized in Mix-of-Show~\cite{gu2023mixofshow}, participants evaluated the generated images based on two key metrics: 1) Text-to-Image Alignment: This assesses how well the textual description matches the generated image. 2) Image-to-Image Alignment: This examines the resemblance between the character in the generated image and the provided character reference image. 

As shown in \cref{fig: user_study}, participants rated each aspect on a scale from 1 to 5, where higher scores denote superior quality. To thoroughly gauge the performance across various multi-object customization scenarios, we included setups involving 2, 3, 4, and 5 customization concepts. The sequence of all image-question pairs was randomized before being presented to 25 different users for evaluation. Each user was tasked with rating a total of 60 questions. The study results are shown in \cref{tab: user study}. Across all scenarios, LoRA-Composer emerged as the preferred choice, receiving the highest score of votes. Notably, our method demonstrated significant strengths, especially in scenarios that required eliminating image-based conditions. These outcomes demonstrate the effectiveness of LoRA-Composer in the generation of multi-concept customized images.

\begin{table}[!t]
    \centering
    \scriptsize
    \resizebox{0.6\linewidth}{!}{
    \begin{tabular}{lcccccc}
    \toprule
        % & \multirow{2}{*}{\textbf{Methods}}
        % & \multirow{2}{*}{\shortstack[c]{Real \\ T2I}}
        % & \multirow{2}{*}{\shortstack[c]{Anime \\ T2I}}
        % & \multirow{2}{*}{\shortstack[c]{Real \\ Condition}}
        % & \multirow{2}{*}{\shortstack[c]{Anime \\ Condition}}
        % & \multirow{2}{*}{\textbf{Mean Result}}
        % \\\\
        \textbf{Method}
        & \textbf{Text-to-Image}
        & \textbf{Image-to-Image}
        \\
\midrule 
 Cones2~\cite{cones2} & 1.99 & 1.25 \\
 Mix-of-Show~\cite{gu2023mixofshow}  & 3.13 & 2.58 \\
 Anydoor~\cite{chen2023anydoor}  & 2.73 & 2.07 \\
 Paint by Example~\cite{yang2023paint}  & 2.19 &  1.53 \\
 \textbf{LoRA-Composer} & \textbf{4.25}  & \textbf{4.02} &   \\
\midrule 
 Mix-of-Show*~\cite{gu2023mixofshow} & 3.84 & 3.28  \\
 \textbf{LoRA-Composer*}  & \textbf{4.23} & \textbf{3.78}\\
\bottomrule
\end{tabular}}
\vspace{2mm}
\caption{User study. The scores reflect user preferences, with higher values indicating better quality. It shows that our approach is favored by users for multi-concept customization, excelling in both image and text alignment. An asterisk \textbf{*} denotes using image-based conditions. The highest scores in each column are marked in \textbf{bold}.}
\vspace{-4mm}
\label{tab: user study}
\end{table}

\begin{figure}[!tb]
  \centering
  \includegraphics[width=\textwidth]{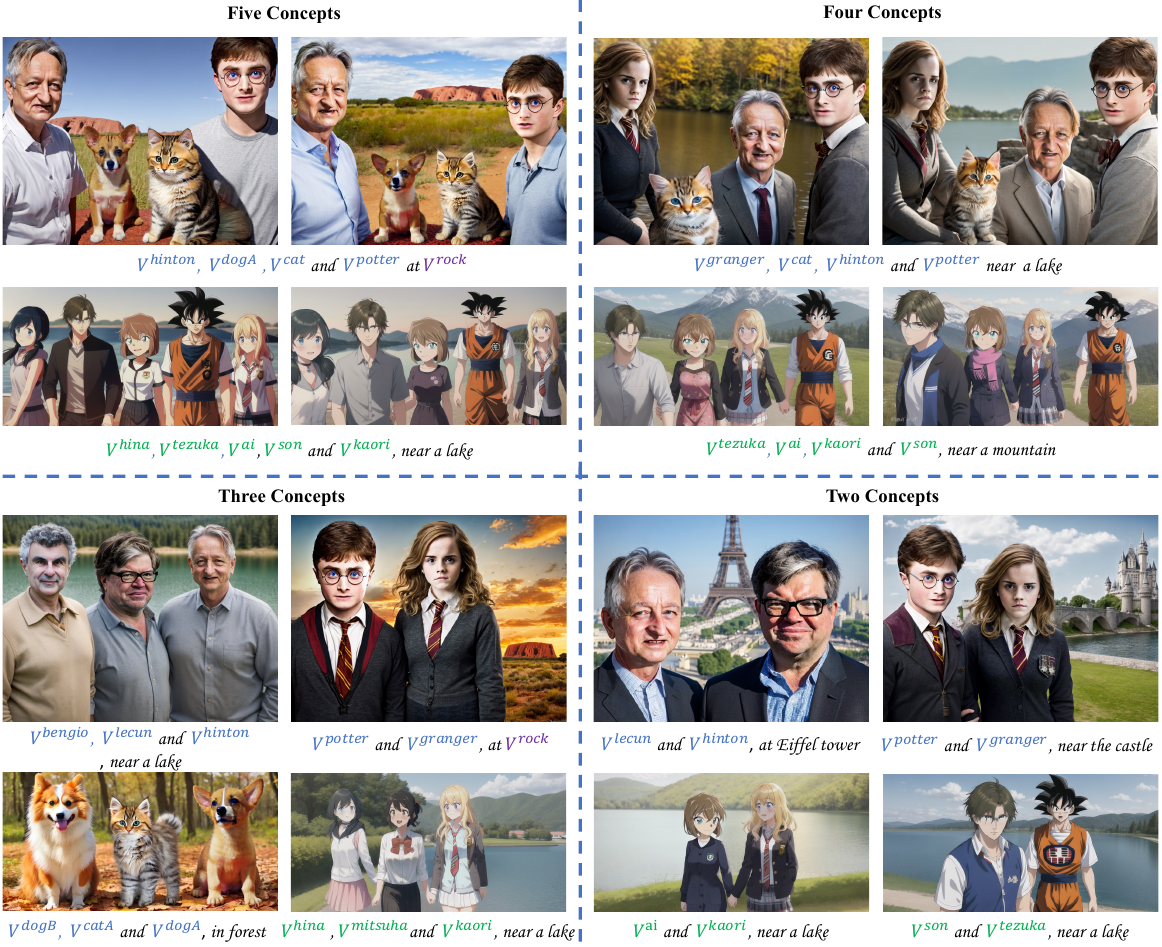}
  \vspace{-2mm}
  \caption{More results of our method in four configurations.}
  \vspace{-2mm}
  \label{fig: more_res}
\end{figure}

\subsection{More Visual Results}
\label{More Visual Results}
In \cref{fig: more_res}, we showcase an extended collection of images produced using our method. This display illustrates the superior flexibility and usability of LoRA-Composer, enabling users to create images under few conditions and utilizing easily accessible LoRA techniques. Additionally, our method's capability to seamlessly blend multiple concepts into high-fidelity images showcases its effective application in multi-concept generation tasks.

\section{Potential Negative Society Impact}
\label{Potential Negative Society Impact}

This project is dedicated to offering the community an advanced tool for multi-concept image customization, empowering users to merge various concepts seamlessly to craft complex visuals. Nonetheless, there's a risk that such a powerful framework could be misused by malicious parties to create deceptive interactions with real-world figures, posing potential harm to the public. 
% This concern transcends our model, affecting the broader field of multi-concept customization. 
To counteract these risks, one potential solution is implementing protective measures akin to those proposed in DUAW~\cite{ye2023duaw}, which introduces a universal adversarial watermark. This watermark is designed to interfere with the variational autoencoder's function, thereby hindering the model's ability to be exploited for malicious customization. 
% Moreover, embedding watermarks into the generated images could serve as a safeguard for copyright, ensuring that the use of such images without appropriate authorization or acknowledgment is deterred. These strategies aim not only to enhance the model's security but also to uphold ethical standards within digital content creation.

\section{Limitation and Future Work}
\label{Limitation}
\begin{figure}[!tb]
  \centering
  \includegraphics[width=\textwidth]{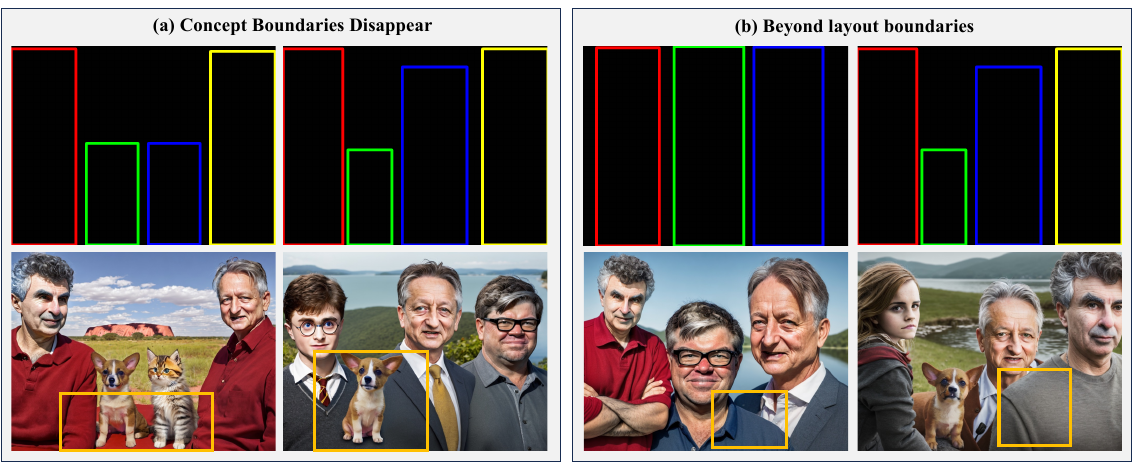}
  \caption{Limitation of LoRA-Composer. (a) Concept boundaries disappear. (b) Beyond layout boundaries.}
    % \vspace{-5mm}
  \label{fig: limitation}
\end{figure}

The first limitation is about disappearing concept boundaries (\cref{fig: limitation}(a)): This issue arises when the space between concepts is too small, causing potential overlap due to down-sampling. Increasing the spacing between concepts can alleviate this problem. 

The second limitation pertains to instances where concepts extend beyond their designated layout boundaries, as shown in \cref{fig: limitation}(b). Occasionally, foreground elements may spill over their intended borders, a consequence of Stable Diffusion's~\cite{rombach2021highresolution} design, which relies on generalized assumptions to generate outcomes. Adopting a more structured layout strategy could potentially mitigate this issue.

The final limitation pertains to inference efficiency. A slight delay occurs due to the need to load various LoRA checkpoints and perform backward computations to update latent representations. This process takes approximately 20-40 seconds per image on a single NVIDIA A100 GPU.

In future work, we aim to enhance the attention mechanism to overcome existing limitations and optimize the IO process to improve inference efficiency.

%%%%%%%%%%%%%%%%%%%%%%%%%%%%%%%%%%%%%%%%%%%%%%%%%%%%%%%%%%%%

\end{document}